%% file: acl_latex.tex
\pdfoutput=1

\documentclass[11pt]{article}

\usepackage[final]{acl}

\usepackage{times}
\usepackage{latexsym} 

\usepackage[T1]{fontenc}

\usepackage[utf8]{inputenc}

\usepackage{microtype}

\usepackage{inconsolata}

\usepackage{graphicx}

\usepackage{tcolorbox}
\tcbuselibrary{breakable}

\usepackage{booktabs}
\usepackage{multirow}
\usepackage{subfigure}
\usepackage{wrapfig}
\usepackage{caption}
\usepackage{amsmath,amssymb}
\newcommand{\MODELNAME}{\textsc{BRIEF}}
\newcommand{\UNIMODELNAME}{\textsc{UniBRIEF}}

\title{\MODELNAME{}: Bridging Retrieval and Inference for Multi-hop Reasoning via Compression}

\author{
Yuankai Li$^1$\thanks{\hspace{0.5mm}Equal contribution.}, Jia-Chen Gu$^2$\footnotemark[1], Di Wu$^2$, Kai-Wei Chang$^2$, Nanyun Peng$^2$  \\
$^1$Fudan University ~ 
$^2$University of California, Los Angeles \\
{\tt yuankaili21@m.fudan.edu.cn}, ~ {\tt gujc@ucla.edu}, \\
{\tt \{diwu,kwchang,violetpeng\}@cs.ucla.edu}
}

\begin{document}
\maketitle
\begin{abstract}
Retrieval-augmented generation (RAG) can supplement large language models (LLMs) by integrating external knowledge. 
However, as the number of retrieved documents increases, the input length to LLMs grows linearly, causing a dramatic increase in latency and a degradation in long-context understanding. 
This is particularly serious for \emph{multi-hop} questions that require a chain of reasoning across documents. 
To accelerate inference, reduce costs, and minimize distractions, this paper presents \MODELNAME{} (\textbf{B}ridging \textbf{R}etrieval and \textbf{I}nference through \textbf{E}vidence \textbf{F}usion), a lightweight approach that performs query-aware multi-hop reasoning by compressing retrieved documents into highly dense textual summaries to integrate into in-context RAG.
To enable learning compression for multi-hop reasoning, we curate synthetic data by extracting atomic \emph{propositions} that encapsulate distinct factoids from the source documents to compose synthetic summaries. 
Based on our synthetic data built entirely by open-source models, \MODELNAME{} generates more concise summaries and enables a range of LLMs to achieve exceptional open-domain question answering (QA) performance.
For example, on HotpotQA, \MODELNAME{} improves the compression rate by 2 times compared to the state-of-the-art baseline, while outperforming it by 3.00\% EM and 4.16\% F1 with Flan-UL2 as the reader model.
It also generates more concise summaries than proprietary GPT-3.5, while demonstrating nearly identical QA performance\footnote{Code and data: \href{https://github.com/JasonForJoy/BRIEF}{https://github.com/JasonForJoy/BRIEF}}. 
\end{abstract}

\input{text/1_intro}

\input{text/3_preliminary}

\input{text/4_method}

\input{text/5_experiment}

\input{text/2_related}

\input{text/6_conclusion}

\section*{Acknowledgement}
This research is based upon work supported by NSF CAREER \#2339766, an Amazon AGI Research Award, and Okawa Foundation Research Grant.
We thank Sara Khosravi, Yufei Tian, Sidi Lu, and UCLA NLP group members for their valuable feedback.

\bibliography{custom}

\clearpage
\appendix
\onecolumn
\input{text/_appendix}

\end{document}

%% file: text/1_intro.tex
\section{Introduction}
Large language models (LLMs)~\citep{DBLP:conf/nips/BrownMRSKDNSSAA20, DBLP:conf/nips/Ouyang0JAWMZASR22, DBLP:journals/corr/abs-2302-13971} are prone to hallucinations~\citep{DBLP:journals/csur/JiLFYSXIBMF23} and are inherently limited by the static nature of their pre-training data. 
Retrieval-augmented generation (RAG)~\citep{DBLP:conf/nips/LewisPPPKGKLYR020} addresses this limitation by integrating external dynamic knowledge.

However, there are still several open challenges for RAG.
First, the input length grows linearly with the number of retrieved documents, leading to substantial increases in latency and computational costs~\citep{DBLP:conf/emnlp/JiangWLYQ23,DBLP:conf/iclr/XuSC24}.
Second, incorporating multiple documents in context is prone to introducing noise, potentially confusing LLMs and degrading their long-context understanding~\citep{DBLP:conf/icml/ShiCMSDCSZ23,DBLP:conf/acl/MallenAZDKH23}.
Third, long-context LLMs struggle with the \emph{lost-in-the-middle} challenge, where they tend to focus on the beginning and end of long contexts while underutilizing critical details buried deep in the middle~\citep{DBLP:conf/iclr/0008PWM0LSBSC24,DBLP:journals/tacl/LiuLHPBPL24}.
These challenges are particularly pronounced for \emph{multi-hop} queries that require reasoning across documents to collect necessary evidence scattered throughout various positions of the documents, which has been overlooked in previous context compression studies~\cite{DBLP:conf/emnlp/JiangWLYQ23,DBLP:conf/emnlp/0001DGL23,DBLP:conf/iclr/XuSC24}.

\begin{figure*}[t]
  \centering
  \includegraphics[width=0.94\textwidth]{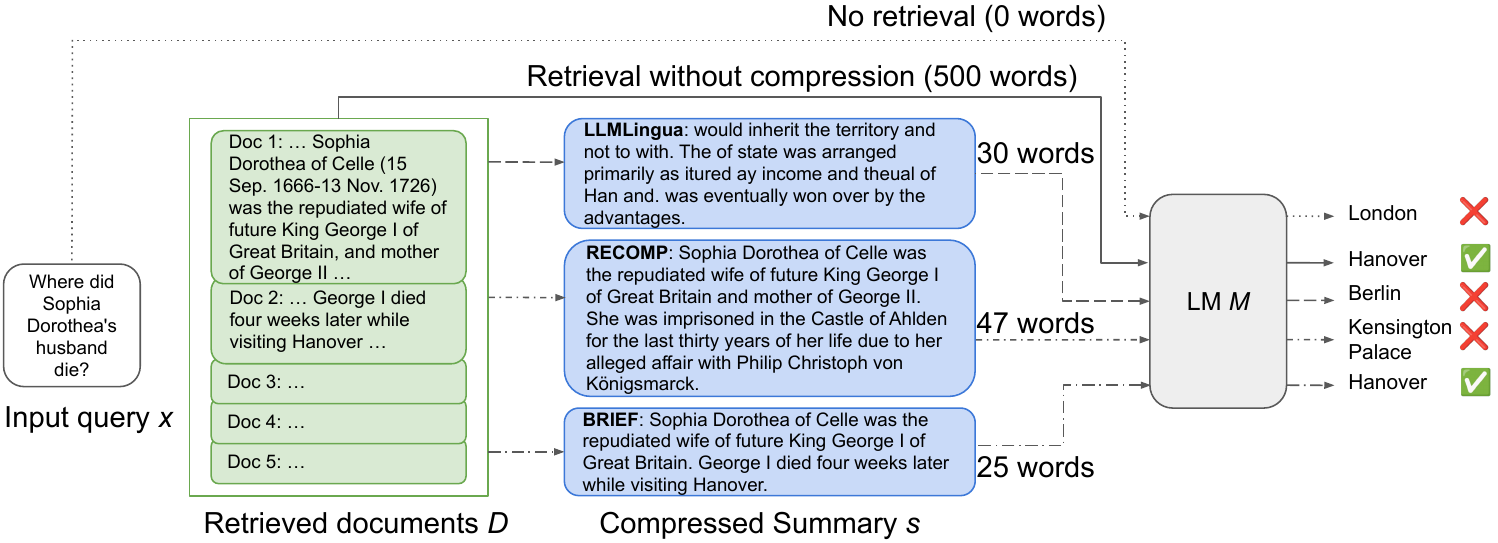}
  \caption{
  A comparison between \MODELNAME{} and previous methods.
  The retrieved documents are compressed into a highly dense textual summary relevant to the query before prepending it as input to an LM.
  LLMLingua~\citep{DBLP:conf/emnlp/JiangWLYQ23} struggles to produce fluent natural language due to its token-level compression. 
  RECOMP~\citep{DBLP:conf/iclr/XuSC24} is limited to collecting evidence in a single logical step, yet it still produces lengthy summaries.
  }
  \label{fig-inference}
  \vspace{-2mm}
\end{figure*}

To accelerate inference, reduce costs, and minimize distractions, we propose \MODELNAME{} (\textbf{B}ridging \textbf{R}etrieval and \textbf{I}nference through \textbf{E}vidence \textbf{F}usion). 
As shown in Figure~\ref{fig-inference}, \MODELNAME{} performs query-aware \emph{multi-hop} reasoning to compress retrieved documents into highly dense textual summaries to integrate into in-context RAG. 
Unlike conventional methods that focus on compression for single-hop questions~\citep{DBLP:conf/iclr/XuSC24,DBLP:conf/acl/CaoCLPHCS24}, \MODELNAME{} is specifically trained to summarize the most pertinent knowledge from multiple documents that is essential for answering multi-hop questions.
Compared to token-, phrase-, or sentence-level compression~\citep{DBLP:conf/emnlp/JiangWLYQ23,DBLP:conf/emnlp/0001DGL23}, the summaries produced by \MODELNAME{} organize and synthesize evidence relevant to the query in a more concise natural language format, making them more effective for use by the follow-up reader LM.
Besides, the lightweight, T5-based~\citep{DBLP:journals/jmlr/RaffelSRLNMZLL20} \MODELNAME{} reduces costs by over 70\% through compression, yet is capable of identifying relevant details in lengthy documents, relieving the burden on LLMs (Section~\ref{sec-analysis}).

The key innovation of \MODELNAME{} lies in its ability to perform document compression and enable a range of LLMs to perform multi-hop reasoning. 
Unlike the state-of-the-art fine-tuned compressor distilled from extreme-scale proprietary LLMs~\citep{DBLP:conf/iclr/XuSC24}, \MODELNAME{} is trained on data synthesized through a pipeline built entirely by open-source models, without relying on any proprietary LLMs or human annotations.
To curate a dataset for compressor training, a synthetic data pipeline as shown in Figure~\ref{fig-pipeline} is designed by extracting atomic \emph{proposition} expressions that encapsulate distinct factoids from the source documents to compose synthetic summaries~\citep{DBLP:conf/emnlp/MinKLLYKIZH23,DBLP:journals/corr/abs-2312-06648}.
The pipeline includes an automatic validation mechanism to filter out spurious multi-hop questions and corresponding summaries, ensuring that only those requiring genuine multi-hop reasoning are retained, ultimately improving the quality and reliability of the synthetic data.
Besides, our approach exhibits impressive awareness of multi-hop reasoning and scalability, offering a data-centric approach to constructing high-quality and cost-effective synthetic data for context compression. 

To measure the effectiveness of the proposed \MODELNAME{}, we evaluate the performance on open-domain question answering (QA) datasets, including HotpotQA~\citep{DBLP:conf/emnlp/Yang0ZBCSM18}, MuSiQue~\citep{DBLP:journals/tacl/TrivediBKS22}, Natural Questions (NQ)~\citep{DBLP:journals/tacl/KwiatkowskiPRCP19}, TriviaQA~\citep{DBLP:conf/acl/JoshiCWZ17}, and the curated multi-hop versions of NQ and TriviaQA. 
Experimental results show that compared to previous long-context compression methods, \MODELNAME{} improves the performance of in-context retrieval augmentation on both multi- and single-hop questions, while prepending significantly fewer words. 
Specifically, \MODELNAME{} compresses documents by 19.19x, significantly higher than RECOMP's 10.02x~\citep{DBLP:conf/iclr/XuSC24}, while still outperforming it by 3.00-point EM and 4.16-point F1 on HotpotQA with Flan-UL2 as the reader language model (LM).
For single-hop questions, \MODELNAME{} compresses documents by 29.76x, significantly higher than RECOMP's 16.23x, while still outperforming it on TriviaQA.
In comparison to proprietary LLM GPT-3.5 as the compressor, \MODELNAME{} demonstrates nearly identical QA performance, while compressing by 19.19x better than GPT-3.5's 14.77x on HotpotQA, and by 17.67x better than GPT-3.5's 11.33x on NQ.

In summary, our contributions in this paper are four-fold:
(1) This study pioneers the exploration of long-context reasoning and compression of RAG for \textit{multi-hop questions}.
(2) A synthetic data pipeline, built entirely by \textit{open-source models}, is designed to enhance the awareness of multi-hop reasoning and scalability due to low cost. 
(3) \MODELNAME{}, trained on the curated dataset, achieves \textit{exceptional QA performance with more concise summaries} compared to proprietary LLM-based compressors. 
(4) We contribute high-quality multi-hop test sets that reveal the limitations of previous compressors, which excel in single-hop settings but fall behind our method in multi-hop settings.

%% file: text/3_preliminary.tex
\section{Preliminaries}

  \subsection{Problem Formulation}
  Given an input sequence $\mathbf{x}$, a target output sequence $\mathbf{y}$, and a set of $N$ retrieved documents $D$ ($[\mathbf{d}_1, \mathbf{d}_2, ..., \mathbf{d}_N]$)\footnote{Improving retriever is not the focus, so we assume a set of retrieved documents are provided following~\citet{DBLP:conf/iclr/XuSC24}.}, 
  \MODELNAME{} compresses retrieved documents $D$ into a summary $\mathbf{s}$ which captures the core information with respect to $\mathbf{x}$ with significantly fewer words. 
  The whole architecture consists of two modules: a compressor $\mathcal{C}$ and an LM $\mathcal{M}$. 
  The compressor $\mathcal{C}$ is trained on the corpora we curated in this work, while the LM $\mathcal{M}$ remains frozen and can be any off-the-shelf LM.
  In this work, we train an encoder-decoder model to serve as an \emph{abstractive} compressor, which takes the input sequence $\mathbf{x}$ and the concatenation of retrieved document set $D$, and outputs a summary $\mathbf{s}$  \citep{DBLP:conf/iclr/XuSC24}. 
  The compressor $\mathcal{C}$ is intentionally designed to be substantially smaller than the LM $\mathcal{M}$, as we aim to reduce the computational cost of encoding a set of lengthy retrieved documents.

  \subsection{Helpfulness Definition} \label{sec-helpful-definition}
  Given a set of retrieved documents, a pressing research challenge is to identify which documents, or more fine-grained segments within them, are the most helpful and can effectively support answering the question. 
  For a question $\mathbf{x}$, the helpfulness of each document $\mathbf{d}_i$ is determined by the LM's end-task performance when the document is prepended.
  Formally, we compare the log likelihood assigned to the target output $\mathbf{y}$ by an LM $\mathcal{M}$ before prepending the document, i.e., $\log p_{\mathcal{M}} (\mathbf{y} | \mathbf{x})$, and after, i.e., $\log p_{\mathcal{M}} (\mathbf{y} | \mathbf{d}_i, \mathbf{x})$. 
  A document is considered helpful for answering the question if the likelihood increases after prepending.
  This approach allows us to filter the retrieved documents $D$ to identify a helpful document subset $\widetilde{D}$. 
  Furthermore, to identify more fine-grained, helpful segments within a document, each document $\mathbf{d}_i \in \widetilde{D}$ is segmented into a set of \emph{atomic expressions} $P_i$ ($[\mathbf{p}_i^1, \mathbf{p}_i^2, ..., \mathbf{p}_i^M]$)\footnote{The atomic expressions can be in any format as designed, e.g., sentence~\citep{DBLP:conf/iclr/XuSC24}, multi-sentence or sub-sentence. In this work, we adopt the granularity of \emph{propositions}~\citep{DBLP:conf/emnlp/MinKLLYKIZH23,DBLP:journals/corr/abs-2312-06648} for the reasons explained in detail in Section~\ref{sec-document-proposition}.}, where $M$ varies across different documents.
  Similarly, we compare the log likelihood assigned to the target output $\mathbf{y}$ by the LM $\mathcal{M}$ before prepending an atomic expression, i.e., $\log p_{\mathcal{M}} (\mathbf{y} | \mathbf{x})$, and after, i.e., $\log p_{\mathcal{M}} (\mathbf{y} | \mathbf{p}_i^j, \mathbf{x})$. 
  An atomic expression is considered helpful for answering the question if the likelihood increases after prepending.
  The helpful atomic expressions are ranked by their associated answer likelihood and the top-\emph{k} are selected as the target summary $\mathbf{s}$.

  \subsection{Parsing a Document to Propositions} \label{sec-document-proposition}
  Multi-hop reasoning is the process of connecting multiple pieces of evidence across different logical steps to reach a conclusion that cannot be derived from any single piece of evidence alone. 
  In this work, each document $\mathbf{d}_i$ is segmented into a set of atomic \emph{propositions}.
  Figure~\ref{fig-proposition-example} in Appendix~\ref{sec-appendix-proposition} presents an example of parsing a document into a set of propositions.
  Document proposition addresses the problem of sentence decontextualization: rewriting a sentence along with its context to make it interpretable out of context, while preserving its original meaning~\citep{DBLP:journals/tacl/ChoiPLKDC21}.
  Propositions encapsulate distinct factoids in a concise and self-contained natural language format, offering improved factoid granularity for information retrieval, fact checking, and open-domain QA~\citep{DBLP:conf/emnlp/MinKLLYKIZH23,DBLP:journals/corr/abs-2312-06648}.  
  Therefore, propositions can serve as foundational units of evidence, and logical connections for answering complex questions can be established by linking information from different propositions. 
  Additionally, proposition-like compression is more compatible across LMs and efficient than token- or sentence-level compression~\citep{DBLP:conf/emnlp/JiangWLYQ23,DBLP:conf/emnlp/0001DGL23}.

%% file: text/4_method.tex
\section{\MODELNAME{}}

\begin{figure*}[t]
  \centering
  \includegraphics[width=0.98\textwidth]{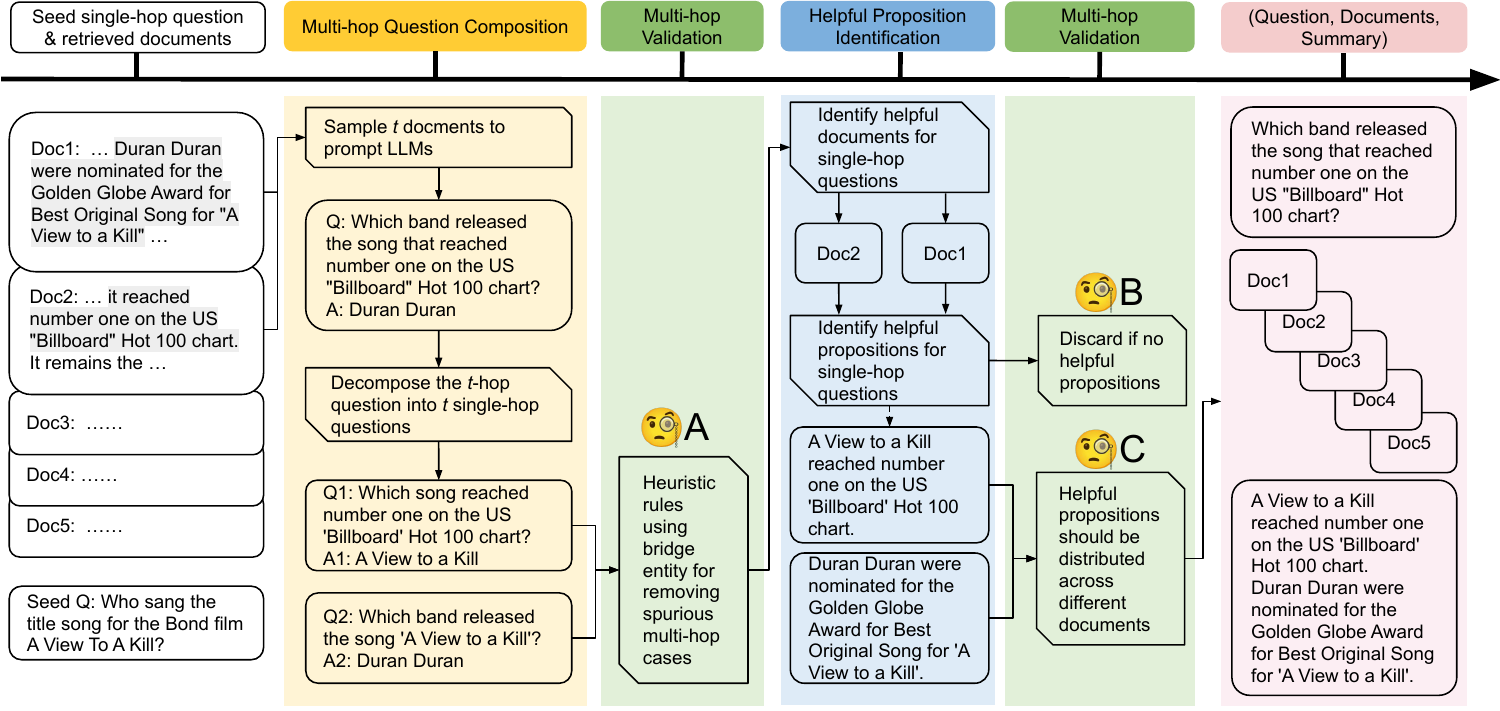}
  \caption{
  An overview of the synthetic data pipeline for training \MODELNAME{}. 
  Starting with a seed single-hop question, the pipeline can generate a \emph{multi-hop} (question, documents, summary) tuple to enhance the awareness of multi-hop reasoning and compression. 
  Meanwhile, it can also generate a \emph{single-hop} tuple through a simplified process by bypassing the \emph{Multi-hop Question Composition} and \emph{Multi-hop Validation} modules.
  }
  \label{fig-pipeline}
  \vspace{-2mm}
\end{figure*}

  \subsection{\MODELNAME{} Inference}
  Figure~\ref{fig-inference} presents an overview of \MODELNAME{} at inference. 
  For every input query $\mathbf{x}$, an off-the-shelf dense passage retriever~\citep{DBLP:conf/emnlp/KarpukhinOMLWEC20,DBLP:journals/tmlr/IzacardCHRBJG22} returns a set of $N$ retrieved documents $D$ ($[\mathbf{d}_1, \mathbf{d}_2, ..., \mathbf{d}_N]$).
  Then, the compressor $\mathcal{C}$ takes as input the concatenation of query $\mathbf{x}$ and retrieved documents $D$, and outputs a summary $\mathbf{s}$.
  If the retrieved documents are considered irrelevant to the input, our compressor can return an empty string, implementing selective retrieval augmentation~\citep{DBLP:conf/iclr/XuSC24}. 
  Finally, the input query $\mathbf{x}$ and the compressed summary $\mathbf{s}$ are fed into an off-the-shelf LM $\mathcal{M}$. 
  Following RECOMP, we include few-shot in-context examples in the prompt. 
  The five in-context examples are randomly sampled from their corresponding training sets.
  Figure~\ref{fig-inference-prompt} in Appendix~\ref{sec-appendix-prompts} presents the detailed inference prompts for each dataset.

  \subsection{Data Collection} \label{sec-data}
  Collecting human annotations to train the compressor $\mathcal{C}$ is quite expensive. 
  \citet{DBLP:journals/corr/abs-2209-12356} and \citet{DBLP:conf/acl/PotluriXC23} have shown that LLMs can generate decent query-focused summaries when carefully prompted.
  Therefore, \citet{DBLP:conf/iclr/XuSC24} distill the summarization knowledge of proprietary LLMs into an in-house abstractive compressor.
  Despite the effectiveness of human-annotated or proprietary LLMs-generated summaries, the data generation process is not reproduce-friendly and impractical to scale up  due to the high costs of human annotation and proprietary LLM invocation. 
  
  Different from all these works, we train the abstractive compressor by designing a synthetic data pipeline as shown in Figure~\ref{fig-pipeline} which is built entirely by open-source models. 
  This pipeline consists of the following modules: \emph{multi-hop question composition}, \emph{multi-hop validation}, and \emph{helpful proposition identification}, focusing on improving compression for multi-hop reasoning.
  The automatic validation mechanism of the pipeline helps filter out spurious multi-hop questions and corresponding summaries, ensuring that only those requiring genuine multi-hop reasoning are retained, ultimately improving the quality and reliability of the synthetic data.

    \subsubsection{Question Composition and Validation} \label{sec-question-composition}
    Answering \emph{multi-hop} questions requires synthesizing information from multiple sources or reasoning across several steps to arrive at an answer. 
    Based on a wealth of available single-hop questions, the pipeline first composes \emph{multi-hop} questions that necessitate a series of inferential or deductive steps to distill and integrate evidence from multiple segments.
    Formally, given a seed single-hop question $\mathbf{x}$ along with its retrieved documents $D$, \emph{t} documents are randomly sampled to derive $\widehat{D}$.
    Since LLMs exhibit impressive abilities to understand instructions and generate fluent questions~\citep{DBLP:conf/emnlp/LiangWZWQL23}, the sampled documents $\widehat{D}$ are fed into open-source LLMs, which are prompted to compose a \emph{t}-hop question $\hat{\mathbf{x}}$ and its answer $\mathbf{\hat{y}}$.
    We utilize the most common relationships identified by \citet{DBLP:conf/emnlp/ZhongWMPC23}, such as \emph{Which continent is \texttt{[ENTITY]} located in?} and \emph{Who is the author of \texttt{[ENTITY]}?}, to aid in discovering connections between separate documents.
    Figure~\ref{fig-generation-multihop-prompt} in Appendix~\ref{sec-appendix-prompts} presents the full relationships and the prompt for multi-hop question composition.

    Although LLMs are effective for multi-hop question composition, the composed questions may appear complex but fail to require genuinely multi-hop reasoning across multiple sources without proper validation.
    Therefore, a robust multi-hop validation mechanism is necessary. 
    Specifically, a composed \emph{t}-hop question $\hat{\mathbf{x}}$ is first decomposed into \emph{t} single-hop questions and their corresponding answers $[(\hat{\mathbf{x}}_1, \hat{\mathbf{y}}_1), ..., (\hat{\mathbf{x}}_t, \hat{\mathbf{y}}_t)]$. 
    By leveraging the concept of \emph{bridge entity}~\citep{DBLP:journals/corr/abs-2401-15391}, some spurious multi-hop questions can be eliminated through heuristic rules which are elaborated in Appendix~\ref{sec-appendix-heuristic}.
    Figure~\ref{fig-decomposition-prompt} in Appendix~\ref{sec-appendix-prompts} presents the prompt for multi-hop question decomposition.
    
    For each single-hop question of $[\hat{\mathbf{x}}_1, ..., \hat{\mathbf{x}}_t]$ decomposed from the remaining multi-hop question $\hat{\mathbf{x}}$, we aim to identify the most helpful propositions $[\hat{\mathbf{p}}_{n_1}^{m_1}, ..., \hat{\mathbf{p}}_{n_t}^{m_t}]$ within the retrieved documents $D$ using the method described in Section~\ref{sec-helpful-definition}, where $\hat{\mathbf{p}}_n^m$ indicates the $m$-th proposition within the $n$-th document. 
    If no helpful propositions can be identified for any single-hop question, this case will be discarded.
    Finally, the helpful propositions should be distributed across different documents, i.e., $[n_1, ..., n_t]$ are distinct from one another. This guarantees that the model must collect relevant evidence from multiple sources. Otherwise, this case will be discarded. 
    These designs coordinate to contribute high-quality multi-hop datasets that reveal the limitations of previous compressors and have been released to support further research.

    \subsubsection{Target Summary} \label{sec-target-summary}
    When validating the multi-hop nature of the synthetic questions, we have identified the helpful propositions $[\hat{\mathbf{p}}_{n_1}^{m_1}, ..., \hat{\mathbf{p}}_{n_t}^{m_t}]$ for the single-hop questions $[\hat{\mathbf{x}}_1, ..., \hat{\mathbf{x}}_t]$ derived from the composed multi-hop question $\hat{\mathbf{x}}$.
    The concatenation of these helpful propositions are considered as the target summary $\hat{\mathbf{s}}$\footnote{We studied merging the list of helpful propositions into a more coherent and natural textual summary by prompting LLMs, without adding or removing any information, but no further improvement was achieved.}.
    So far, we have curated the \emph{multi-hop} training data in the form of $(\hat{\mathbf{x}}, D, \hat{\mathbf{s}}) \sim \mathcal{D}_{comp}$.  

    To allow for the compression of retrieved documents for questions of varying complexity using a unified compressor, we also incorporate training data for handling \emph{single-hop} queries that require straightforward answers derived from a single information source.     
    Their target summaries are defined as the most pertinent propositions using the method described in Section~\ref{sec-helpful-definition}.
    Both single- and multi-hop (question, documents, summary) tuples work in tandem to create a robust dataset $\mathcal{D}_{comp}$ capable of training models for diverse levels of query complexity, enhancing their ability to tackle both simple and intricate query tasks.
    
    It is notable that if none of the documents in $D$ are considered helpful, i.e., $\widetilde{D} = \emptyset$, the target summary is set to an empty string to facilitate selective retrieval augmentation.
    A portion of this data is incorporated into the training set $\mathcal{D}_{comp}$ to enable the compressor $\mathcal{C}$ to generate an empty summary when the retrieved documents are irrelevant or unhelpful for answering the question, thereby mitigating the risk of prepending irrelevant information.

    \input{table/tab_statistics}

  \subsection{\MODELNAME{} Training}
  The curated dataset $\mathcal{D}_{comp}$ is utilized to fine-tune the compressor $\mathcal{C}$, a T5-large model (770M)~\citep{DBLP:journals/jmlr/RaffelSRLNMZLL20} for a fair comparison with RECOMP.
  T5 is pre-trained on summarization datasets~\citep{DBLP:conf/nips/HermannKGEKSB15} and is commonly used in prior studies~\citep{DBLP:journals/corr/abs-2312-06648,DBLP:conf/iclr/XuSC24,DBLP:journals/corr/abs-2404-10774}. 
  The fine-tuning process follows the standard next token objective, formulated as:
  \begin{equation}
    \max_{\mathcal{C}}\mathbb{E}_{(\mathbf{x}, D, \mathbf{s}) \sim \mathcal{D}_{comp}} \log p_{\mathcal{C}} (\mathbf{s} | \mathbf{x}, D). 
    \label{equ-objective}
  \end{equation}

%% file: table/tab_statistics.tex
\begin{table}[t]
  \centering
  \setlength{\tabcolsep}{2.0pt}
  \small 
  \begin{tabular}{lcrc}
  \toprule
                            &       & \# samples & \# words/summary \\
  \midrule
  \multirow{3}{*}{(MultiHop-)TriviaQA} & 1-hop &  29,372  & 15.85 \\
                                       & 2-hop &   5,524  & 25.76 \\
                                       & 3-hop &     251  & 27.20 \\
  \midrule
  \multirow{3}{*}{(MultiHop-)NQ}       & 1-hop &  24,294  & 26.55 \\
                                       & 2-hop &   5,138  & 27.74 \\
                                       & 3-hop &     559  & 28.92 \\
  \midrule
  \multirow{3}{*}{HotpotQA}            & 2-hop &  13,761  & 21.68 \\
                                       & 3-hop &   4,072  & 22.86 \\
                                       & 4-hop &   1,178  & 24.25 \\
  \midrule
  \multirow{3}{*}{MuSiQue}             & 2-hop &   3,673  & 25.38 \\
                                       & 3-hop &     455  & 35.39 \\
                                       & 4-hop &      62  & 47.89 \\
  \bottomrule
  \end{tabular}  
  \caption{The statistics of the curated $\mathcal{D}_{comp}$.}
  \vspace{-2mm}
  \label{tab-statistics}
\end{table}

%% file: text/5_experiment.tex
\section{Experiments}

  \subsection{Experimental Settings}

    \input{table/tab_result_multihop}
  
    \paragraph{Datasets}
    We evaluated \MODELNAME{} on the following datasets: \textbf{HotpotQA}~\citep{DBLP:conf/emnlp/Yang0ZBCSM18}, \textbf{MuSiQue}~\citep{DBLP:journals/tacl/TrivediBKS22}, \textbf{Natural Questions (NQ)}~\citep{DBLP:journals/tacl/KwiatkowskiPRCP19}, and \textbf{TriviaQA}~\citep{DBLP:conf/acl/JoshiCWZ17}. 
    Notably, the first two datasets primarily consist of multi-hop questions, whereas the latter two are mainly composed of single-hop questions.
    Especially for TriviaQA and NQ, we have curated high-quality multi-hop versions using our proposed synthetic data pipeline, named \textbf{MultiHop-TriviaQA} and \textbf{MultiHop-NQ}, which were then randomly split into train/dev/test sets with an 80\%/10\%/10\% distribution.
    We reported results on the test sets of MultiHop-TriviaQA and MultiHop-NQ, and the dev set of MuSiQue.
    Following~\citet{DBLP:conf/iclr/XuSC24}, we reported results on the test set of TriviaQA, the dev set of NQ, and a randomly sampled subset of 500 examples from the dev set of HotpotQA. 

    Table~\ref{tab-statistics} presents the statistics of the curated dataset $\mathcal{D}_{comp}$.
    As the HotpotQA and MuSiQue datasets are inherently multi-hop, we composed multi-hop questions exclusively for the NQ and TriviaQA datasets. 
    The maximum number of hops of question composition for NQ and TriviaQA was set to three, i.e., the resulting datasets, Multihop-NQ and Multihop-TriviaQA, comprise questions with a maximum of three hops.

    \vspace{-1mm}
    \paragraph{Metrics}
    \textbf{Exact match (EM)} and \textbf{F1} of answer strings were reported for QA performance.
    The \textbf{compression rate} was also reported, defined as the ratio of the number of words in the retrieved documents $D$ before compression to the number of words in the compressed summary $\mathbf{s}$ after compression.
    A higher compression rate indicates a shorter summary.    

    \vspace{-1mm}
    \paragraph{Baselines}
    We compared \MODELNAME{} with:
    (1) The off-the-shelf \textbf{T5-large}~\citep{DBLP:journals/jmlr/RaffelSRLNMZLL20}. 
    (2) \textbf{LLMLingua}~\citep{DBLP:conf/emnlp/JiangWLYQ23} performs both coarse-grained, demonstration-level compression and fine-grained, token-level compression, leveraging the perplexity of each demonstration or token calculated by a causal LM. 
    (3) \textbf{Selective Context}~\citep{DBLP:conf/emnlp/0001DGL23} employs a causal LM to calculate self-information for each token, merges tokens into lexical units, and eliminates content that is deemed least necessary.
    (4) \textbf{RECOMP}~\citep{DBLP:conf/iclr/XuSC24} distills the summarization knowledge of proprietary LLMs (\texttt{gpt-3.5-turbo}) into an abstractive compressor T5-large.
    (5) \textbf{GPT-3.5} (\texttt{gpt-3.5-turbo}) is prompted to summarize the documents with respect to the question. 
    In addition, the results of not prepending any documents (\textbf{No documents}) and prepending the Top-5 retrieved documents without compression (\textbf{Top-5 documents}) were also provided for reference.

    \vspace{-1mm}
    \paragraph{Implementation}
    T5-large (770M)~\citep{DBLP:journals/jmlr/RaffelSRLNMZLL20} and Flan-UL2 (20B)~\citep{DBLP:journals/jmlr/ChungHLZTFL00BW24} were adopted as the compressor $\mathcal{C}$ and LM $\mathcal{M}$ respectively following ~\citet{DBLP:conf/iclr/XuSC24} to ensure all results were comparable. 
    Contriever~\citep{DBLP:journals/tmlr/IzacardCHRBJG22} trained on MS MARCO dataset was adopted as a retriever on Wikipedia corpus from Dec. 20, 2018 for all datasets. 
    The articles were truncated into non-overlapping documents of 100 words.
    We prompted \texttt{Llama-3-70B-Instruct}~\citep{llama3modelcard} for multi-hop question composition and decomposition. 
    For each seed question, we repeated the sampling three times for data augmentation.
    We adopted the propositionizer released by \citet{DBLP:journals/corr/abs-2312-06648} for segmenting documents, which is a Flan-T5-large model fine-tuned on the curated document-to-propositions data\footnote{https://github.com/chentong0/factoid-wiki}.
    Refer to Appendix~\ref{sec-appendix-training} for more details.

  \subsection{Experimental Results}
    
    \paragraph{Multi-hop Results}
    Table~\ref{tab-result-multihop} presents the results of open-domain \emph{multi-hop} QA with Flan-UL2 as the LM $\mathcal{M}$.
    \MODELNAME{} demonstrates promising multi-hop performance in both QA and document compression.
    Specifically, \MODELNAME{} achieves a compression rate of 19.19x, with only a 1.60-point decrease in EM and a 1.83-point decrease in F1 compared to prepending full documents on HotpotQA.
    Compared to RECOMP, \MODELNAME{} compresses by higher 19.19x than its 10.02x, while still outperforming it by 3.00-point EM and 4.16-point F1 on HotpotQA.
    On MultiHop-NQ, we observed a similar trend, with \MODELNAME{}'s higher 16.85x than RECOMP's 10.84x, while outperforming RECOMP by 3.78-point EM and 4.39-point F1.
    Compared to the proprietary LLM GPT-3.5, \MODELNAME{} achieves higher compression rates while delivering competitive QA performance.
    Take the results on HotpotQA as an example, GPT-3.5 achieves a compression rate of 14.77x, and QA performance of 31.60\% EM and 42.65\% F1. While \MODELNAME{} achieves higher 19.19x and can still deliver nearly similar QA results of 31.20\% EM and 42.07\% F1 performance. 

    \input{table/tab_result_singlehop}

    \vspace{-1mm}
    \paragraph{Single-hop Results}
    Table~\ref{tab-result-singlehop} presents the results of open-domain \emph{single-hop} QA with Flan-UL2 as the LM $\mathcal{M}$. 
    \MODELNAME{} also demonstrates promising performance for single-hop questions. 
    Specifically, \MODELNAME{} achieves a compression rate of 29.76x, with only a 2.55-point decrease in EM and a 3.49-point decrease in F1 compared to prepending full documents on TriviaQA.
    On NQ, we observed a similar trend, with a compression rate of 17.67x, resulting in only a 2.99-point decrease in EM and a 3.28-point decrease in F1. 
    Compared to RECOMP, \MODELNAME{} compresses by higher 29.76x than its 16.23x, while still outperforming RECOMP on TriviaQA.
    Compared to GPT-3.5, \MODELNAME{} achieves competitive QA performance, while its compression rate of 17.67x significantly outperforms GPT-3.5's 11.33x.

    \vspace{-1mm}
    \paragraph{Discussion on the Tradeoff between Performance and Lantecy}
    One main focus of this work is to examine the critical tradeoff between effectiveness and efficiency in the RAG setting, recognizing that, in many real-world applications, efficiency is as vital as effectiveness. Understanding this tradeoff is essential for optimizing RAG models to meet diverse operational and practical constraints. We emphasize that there are scenarios where \emph{computational resources are a concern}, or when there are \emph{stringent requirements for real-time reasoning speed}. In these scenarios, slightly reduced accuracy may be an acceptable compromise for a model that operates faster, uses fewer resources, and can be deployed more broadly. 
    The key advantage of our method lies in that it gives a better tradeoff between effectiveness and efficiency compared to previous work. 
    It can achieve decent, if not completely comparable, QA performance as non-compressed models while being highly efficient. 
    By compression, our approach reduces the need for processing large amounts of text while still maintaining the core semantics relevant to the query. 
    This leads to faster processing times and lower resource consumption, which is crucial in real-world applications where scalability and speed are essential.

  \subsection{Analysis} \label{sec-analysis}

\begin{figure}[t]
  \centering
  \subfigure{
  \includegraphics[width=3.7cm]{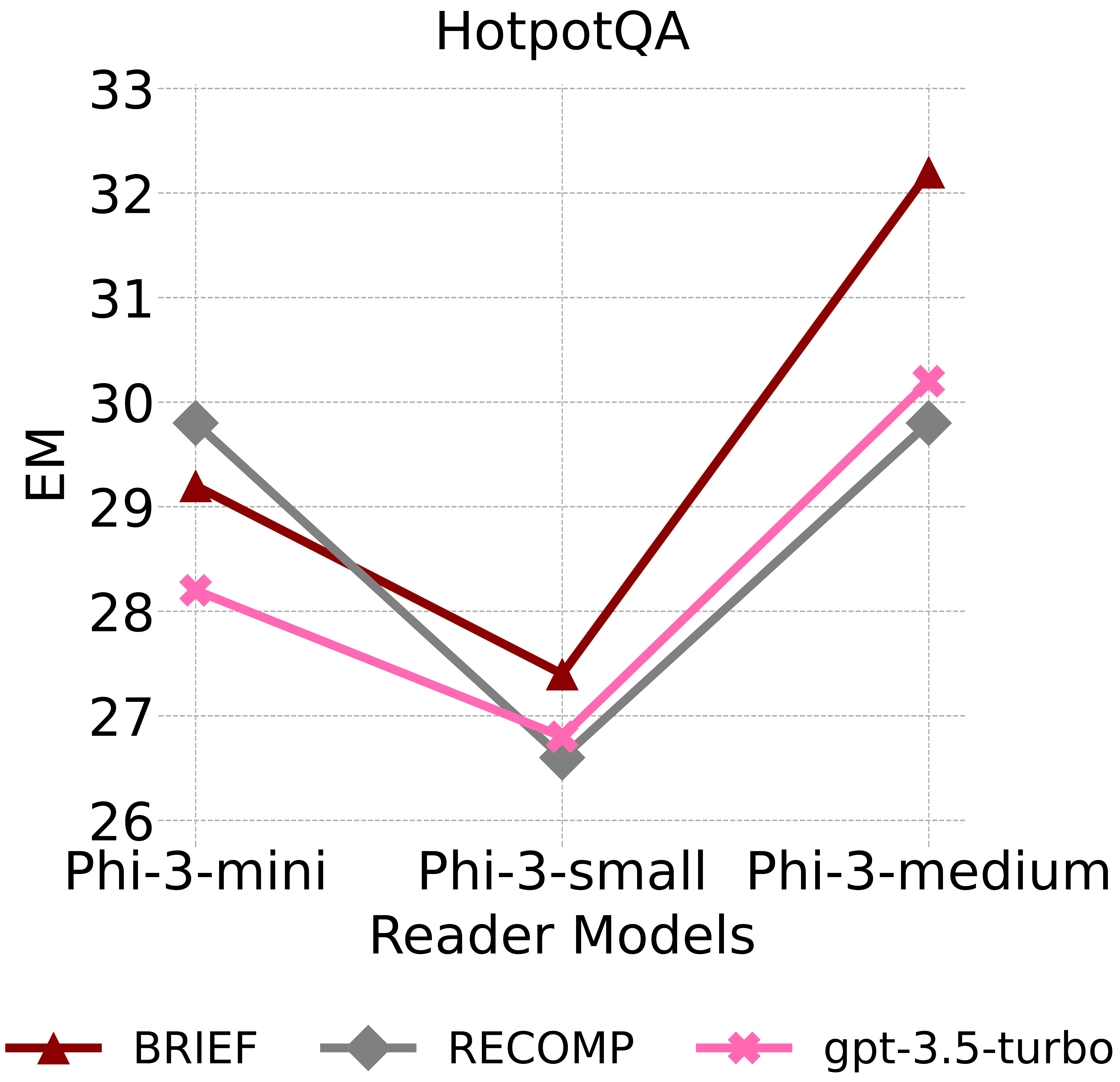}}
  \subfigure{
  \includegraphics[width=3.7cm]{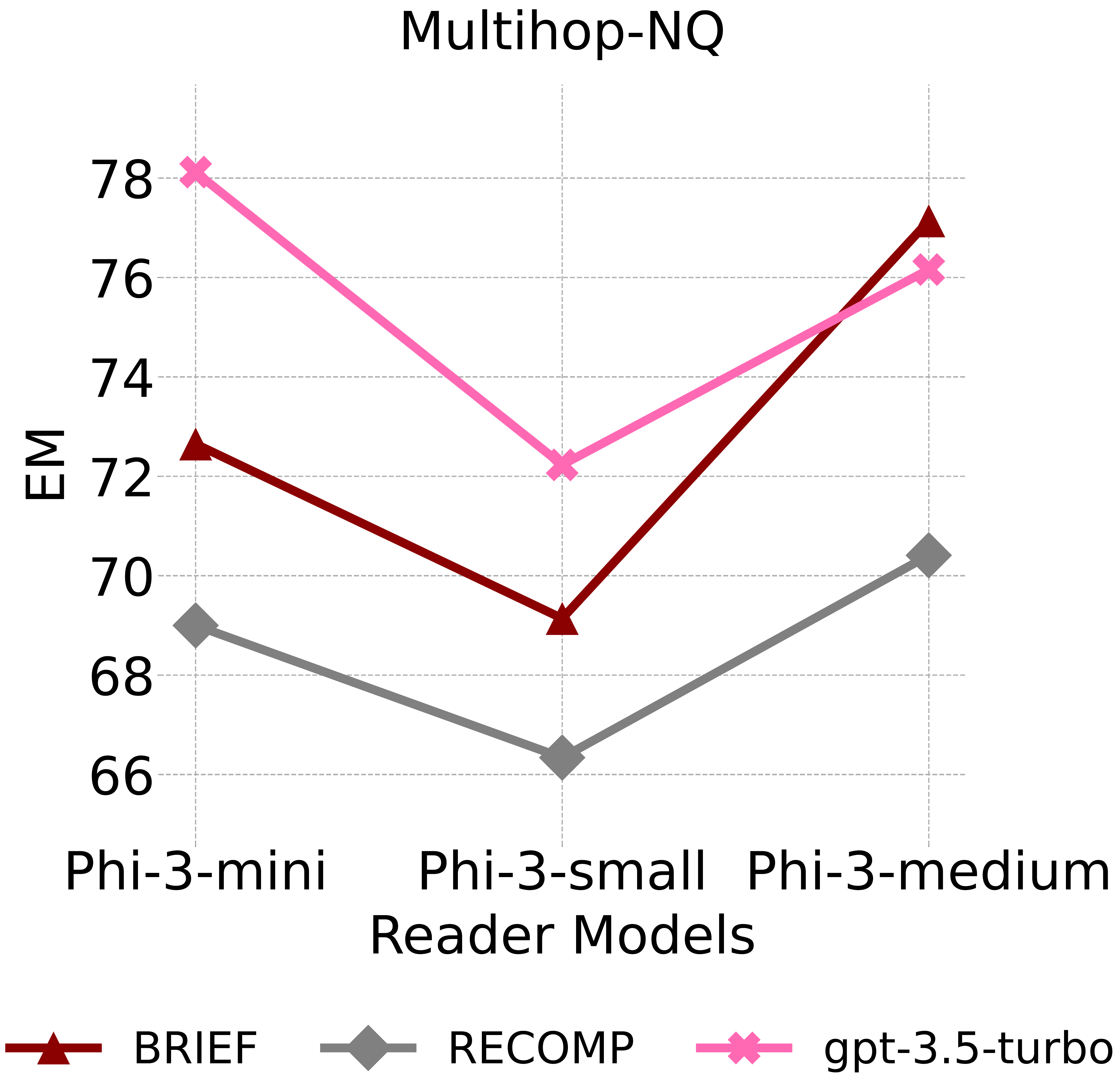}}
  \caption{The transfer ability of compressed summaries across LMs. We selected models from the same family to avoid model selection bias.}
  \vspace{-2mm}
  \label{fig-analysis-lm-transfer}
\end{figure}
    
    \paragraph{The transfer ability of compressed summaries across LMs}
    This ability involves evaluating how well a compressed summary can maintain the core semantics relevant to the query, while also using an expression format that is compatible with a wider range of LMs.
    Therefore, the same sets of compressed summaries were fed into LMs of varying sizes.
    We selected models from the same family to avoid model selection bias, including Phi-3-mini-instruct (3.8B), Phi-3-small-instruct (7B), and Phi-3-medium-instruct (14B)~\citep{DBLP:journals/corr/abs-2404-14219}.
    Figure~\ref{fig-analysis-lm-transfer} presents the QA-compatible performance. 
    It is surprising that Phi-3-mini is a small yet highly capable LM for this task\footnote{Flan-UL2 was chosen as the LM in Table~\ref{tab-result-multihop} and~\ref{tab-result-singlehop} to align with the setting used in RECOMP. Results show that, despite its smaller size, Phi-3-mini is highly capable for this task.
    It is intriguing to see the performance drop in Phi-3 small. We speculate that it may be related to the nature of the Phi-3 series itself. In Phi-3 report~\citep{DBLP:journals/corr/abs-2404-14219}, Phi-3-small (58.1) underperforms Phi-3-mini (64.0) on TriviaQA. }.
    Since our compression takes the form of propositions, it is more interpretable and transfers better across LMs compared to RECOMP and GPT-3.5.
    In comparison to RECOMP and GPT-3.5 on all multi-hop datasets, the performance of \MODELNAME{} drops more slightly when transferring from Phi-3-mini to Phi-3-small, and enlarges more from Phi-3-small to Phi-3-medium.
    These results implied the robustness and consistency of the compressed summaries generated by \MODELNAME{}.

\begin{figure}[t]
  \centering
  \subfigure{
  \includegraphics[width=3.7cm]{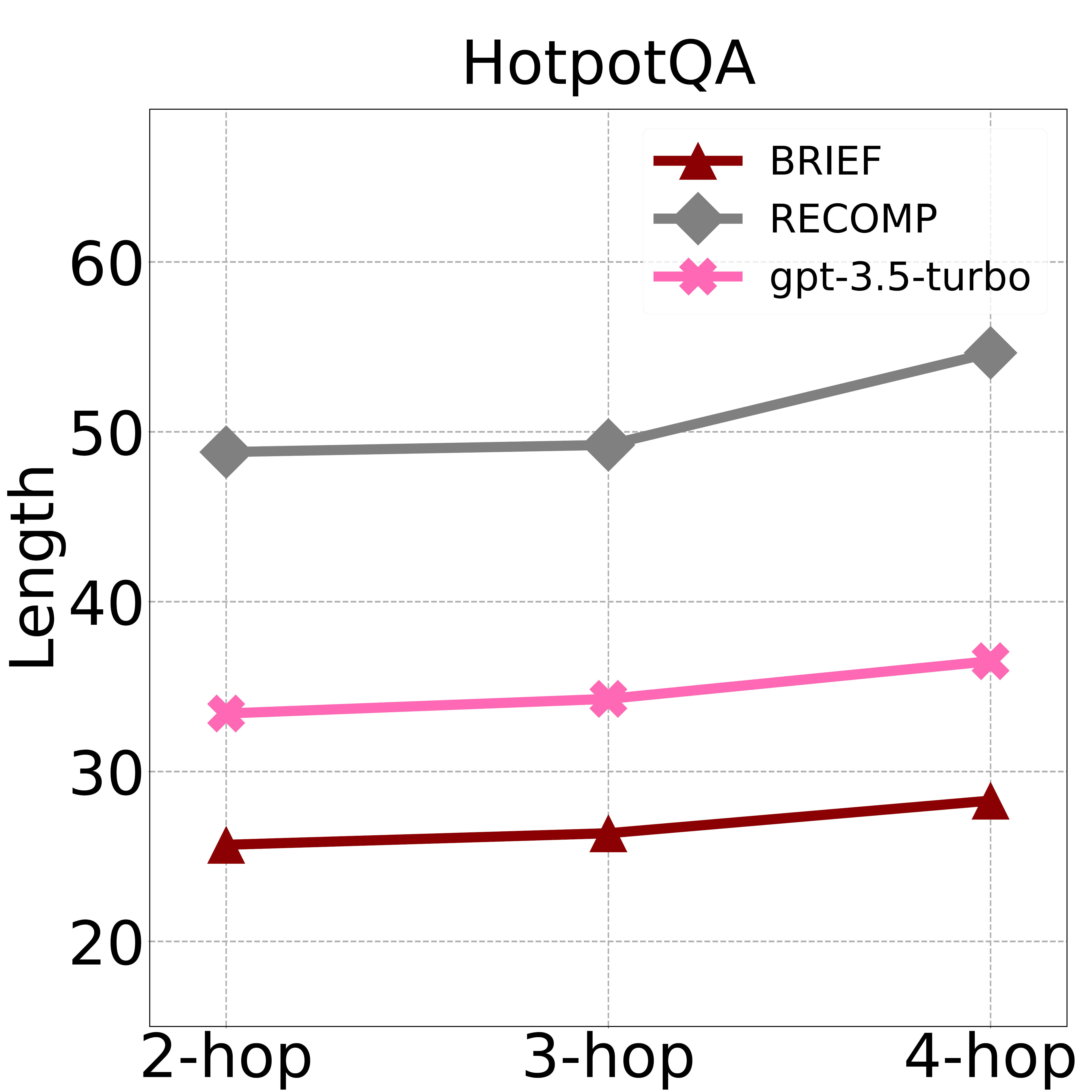}}
  \subfigure{
  \includegraphics[width=3.7cm]{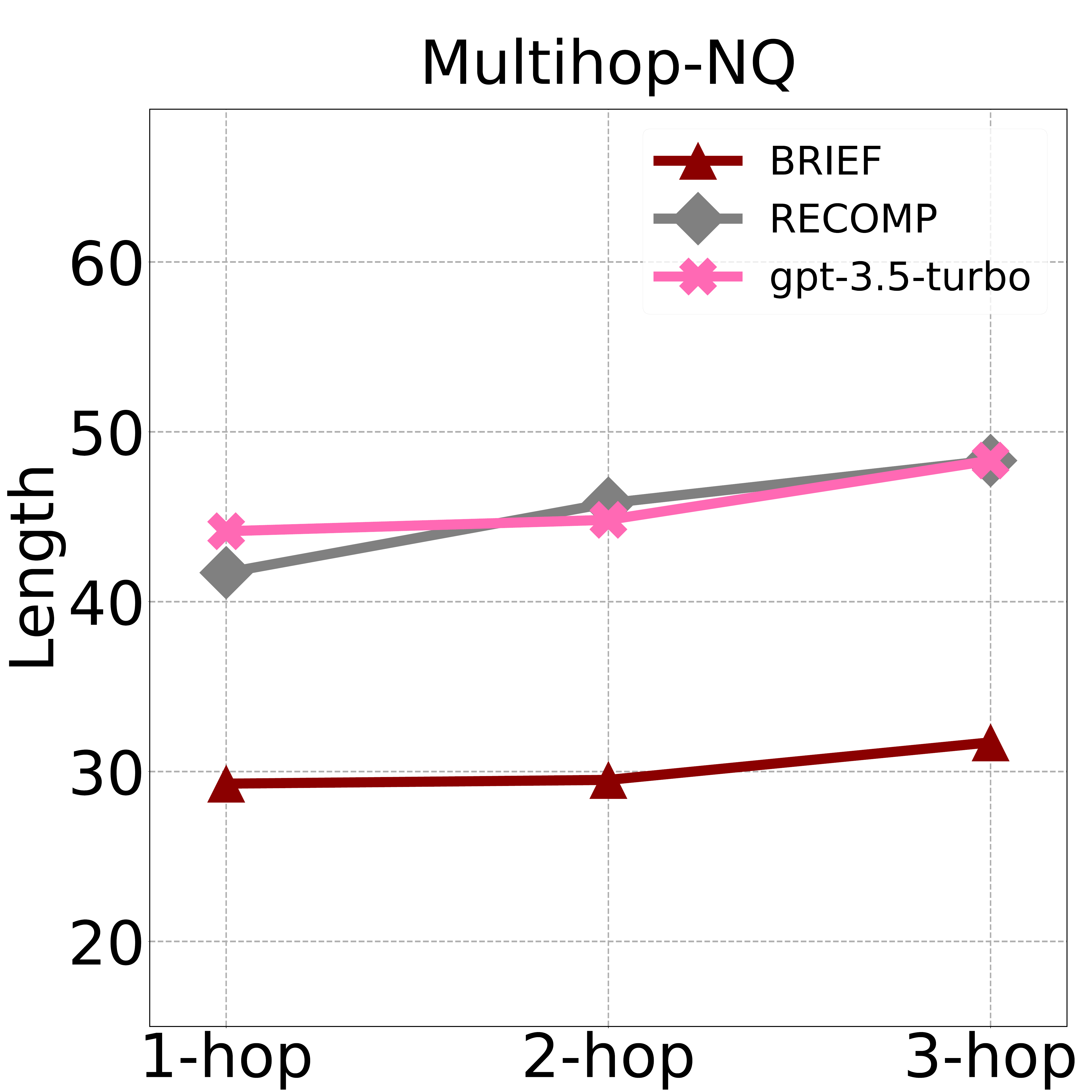}}
  \caption{The length change of compressed summaries with respect to the multi-hop nature of questions.}
  \vspace{-2mm}
  \label{fig-analysis-length-sensitivity}
\end{figure}

    \vspace{-1mm}
    \paragraph{The sensitivity of summary length to multi-hop nature of questions}
    The compressed summaries in response to complex questions tend to be longer, as they need to include more intermediate knowledge to enable adequate reasoning. 
    Therefore, the variation in summary length regarding question complexity can, to some extent, reflect the compressor's sensitivity to that complexity.
    The results is shown in Figure~\ref{fig-analysis-length-sensitivity}.
    As there is no established ground truth for the length of compressed summaries for each question, the results from GPT-3.5 were used as the reference oracle.
    It is important to focus more on the trend of changes across the question hops rather than on the absolute summary length.
    The results indicate that \MODELNAME{} consistently aligns with GPT-3.5 in terms of the sensitivity to the multi-hop nature of questions while generating more concise summaries. 
    This alignment suggests that \MODELNAME{} effectively understands the complexity of questions and adaptively collects the necessary evidence based on specific demands to formulate a complete and accurate summary for answering this question.

\begin{figure}[t]
  \centering
  \subfigure{
  \includegraphics[width=3.7cm]{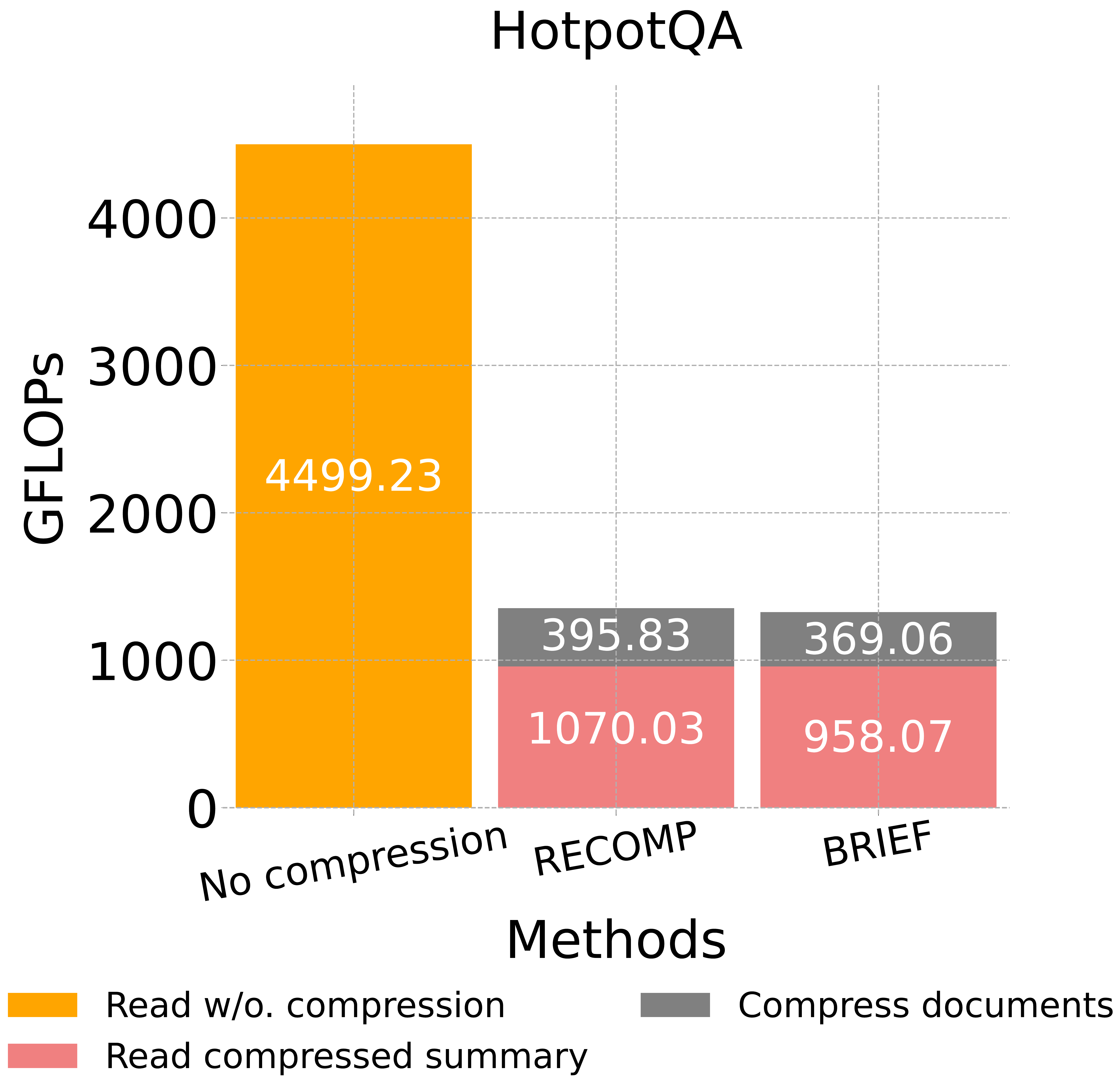}}
  \subfigure{
  \includegraphics[width=3.7cm]{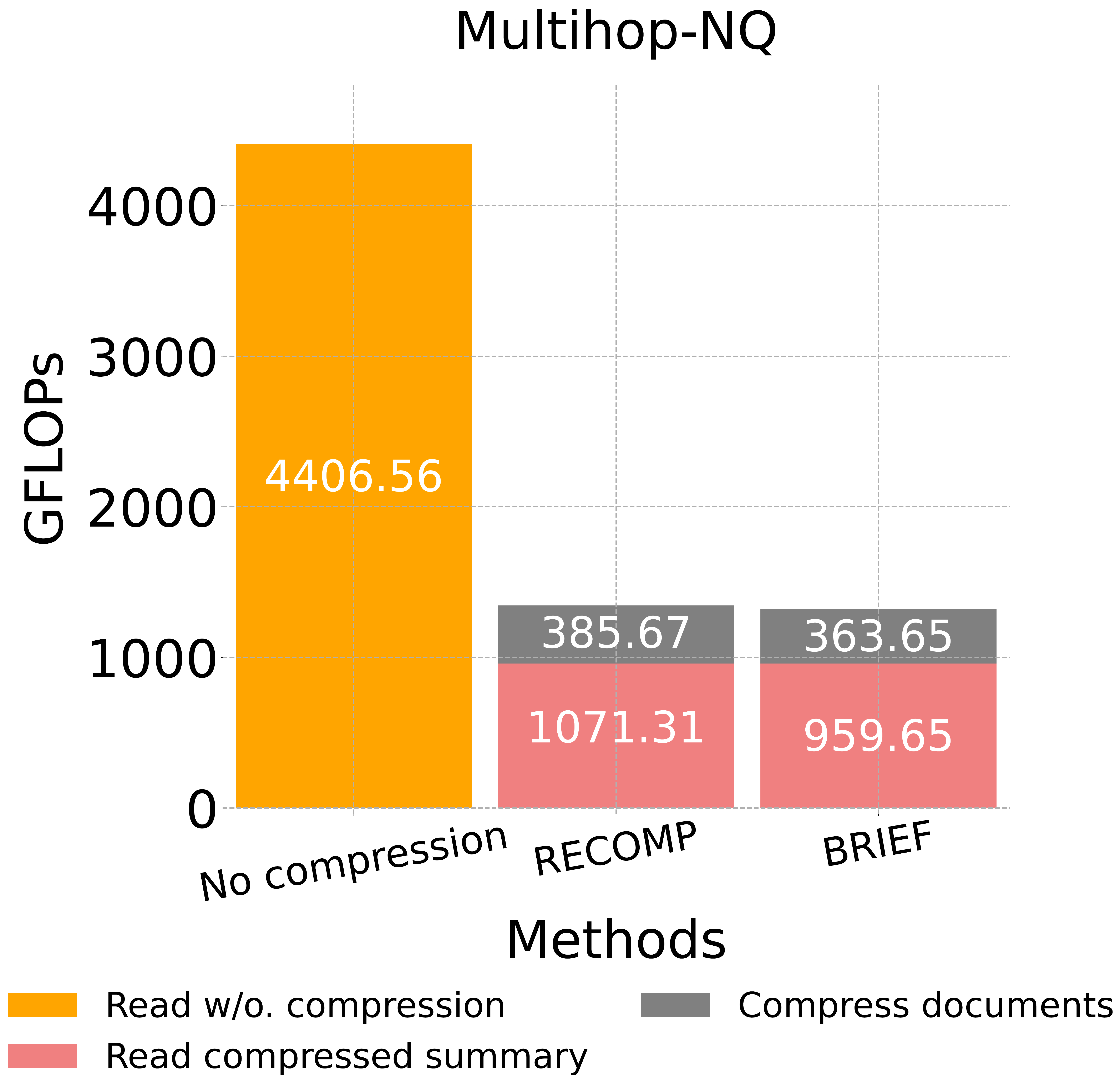}}
  \caption{The comparison of GFLOPs consumption when processing the top-5 documents with or without compression, using Flan-UL2 as the LM.}
  \vspace{-2mm}
  \label{fig-analysis-flops}
\end{figure}

    \vspace{-1mm}
    \paragraph{The improvement of latency in terms of the overall computational overhead}
    The comparison of GFLOPs consumption of processing retrieved documents is shown in Figure~\ref{fig-analysis-flops}.
    The profiler provided by Accelerate to count flops was adopted\footnote{\url{https://huggingface.co/docs/accelerate/usage\_guides/profiler}}.
    Specifically, when employing \MODELNAME{} for compression, the number of GFLOPs required to process compressed documents is significantly reduced compared to the amount required when using Flan-UL2 alone on the original, uncompressed set of top-5 documents. 
    The total amount of computation is reduced to less than 30\% of what it was before compression, which is significantly lower.
    This reduction in GFLOPs highlights \MODELNAME{}'s potential to optimize inference, especially for large-scale document retrieval and processing, by enabling the LM to focus on compressed, more relevant information while maintaining comparable accuracy.

\begin{figure}[t]
  \centering
  \subfigure{
  \includegraphics[width=3.7cm]{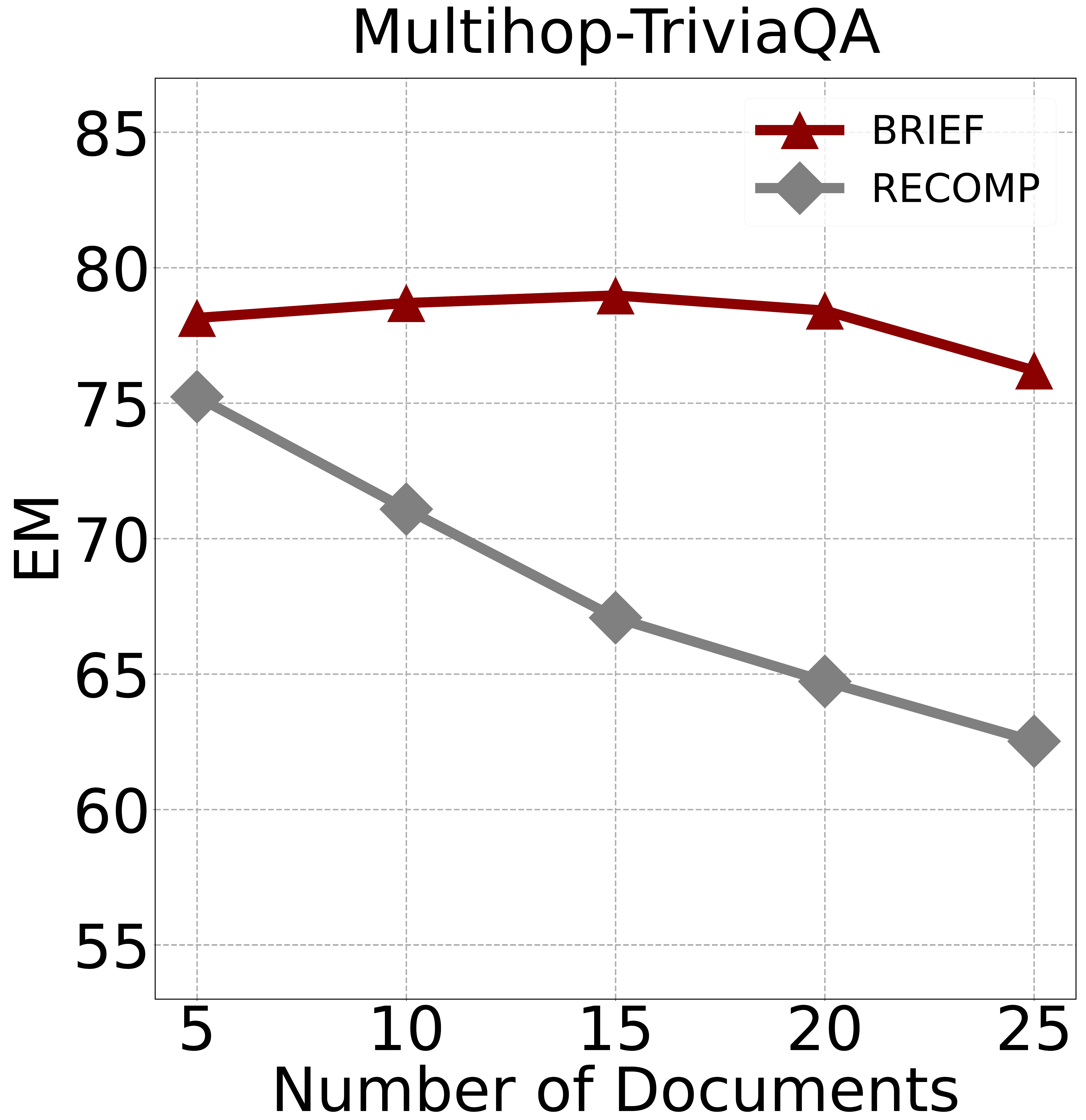}}
  \subfigure{
  \includegraphics[width=3.7cm]{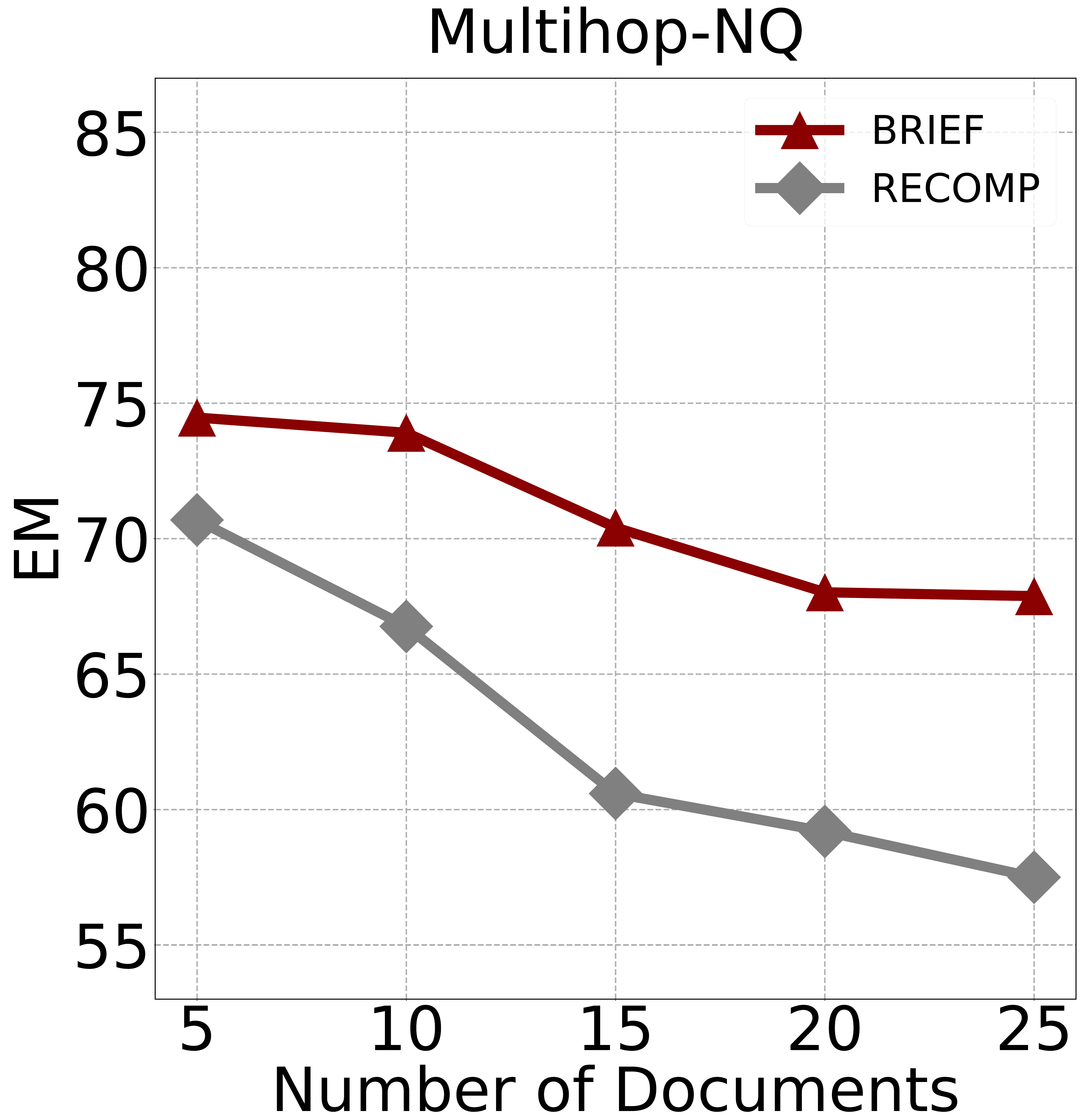}}
  \caption{\emph{Multi-hop} QA results under compression of longer documents, using Flan-UL2 as the LM.}
  \vspace{-2mm}
  \label{fig-analysis-scale}
\end{figure}
    
    \vspace{-1mm}
    \paragraph{The scalability to compress longer documents}
    The maximum sequence length of the T5-based compressor is 512 tokens, which makes it challenging to compress longer contexts that exceed this limit.
    We further explored whether the proposed compressors could be effectively applied to more complex scenarios, particularly those involving documents whose lengths are an order of magnitude longer.
    A preliminary study was conducted by expanding the scope of retrieved documents from the top-5 to the top-25.
    To avoid document position bias, these documents were shuffled and uniformly divided into document chunks, each containing five documents.
    Each chunk was then compressed using the trained compressor according to standard procedures.
    Finally, the compressed results of each chunk were concatenated to produce the overall compressed summary.
    The results are shown in Figure~\ref{fig-analysis-scale}.
    \MODELNAME{} demonstrates better scalability in scenarios where the document length is significantly longer.
    \MODELNAME{} is relatively stable, while RECOMP shows significant performance degradation.
    This result suggests that RECOMP has a limited ability to identify relevant evidence within a longer context containing more distracting information. 
    Overall, our findings suggest that \MODELNAME{} has the potential to be extended but still requires further investigation for compressing longer contexts, which will be explored in future.

%% file: table/tab_result_multihop.tex
\begin{table*}[t]
  \centering
  \small
  \setlength{\tabcolsep}{1.7mm}
  \begin{tabular}{lcccccccccccc}
  \toprule
                        & \multicolumn{3}{c}{HotpotQA}& \multicolumn{3}{c}{MuSiQue}&  \multicolumn{3}{c}{MultiHop-NQ} &  \multicolumn{3}{c}{MultiHop-TriviaQA}      \\
    Method              &  Rate  &  EM   &  F1   &  Rate  &   EM  &   F1  &  Rate  &  EM    &  F1   &  Rate  &  EM   &  F1    \\
  \midrule
    No documents        &  -     & 17.80 & 26.10 &  -     &  5.63 & 14.58 &  -     & 29.03  & 33.44 &  -     & 27.25 & 29.44  \\
    Top-5 documents     &  1x    & 32.80 & 43.90 &  1x    & 35.48 & 45.26 &  1x    & 81.21  & 84.82 &  1x    & 87.69 & 89.88  \\
  \midrule
    T5                  &  9.91x & 26.80 & 36.11 & 10.18x & 18.76 & 28.86 &  9.80x & 57.92  & 61.91 &  9.94x & 62.66 & 65.17  \\
    GPT-3.5             & \underline{14.77x} & \textbf{31.60} & \textbf{42.65} & \underline{12.33x} & \textbf{32.06} & \textbf{43.77} & \underline{11.09x} & \textbf{77.14}  & \textbf{81.17} & 11.44x & \textbf{82.02} & \textbf{84.25}  \\
    Selective Context   &  4.39x & 23.80 & 32.96 &  4.43x & 12.74 & 21.77 &  4.59x & 47.27  & 52.53 &  4.44x & 51.18 & 53.89  \\
    LLMLingua           &  5.19x & 20.20 & 29.53 &  4.43x &  8.50 & 17.74 &  4.54x & 43.48  & 48.79 &  4.58x & 39.42 & 45.03  \\
    RECOMP              & 10.02x & 28.20 & 37.91 &  8.65x & 24.45 & 33.95 & 10.84x & 70.69  & 73.89 & \underline{16.47x} & 75.24 & 77.78  \\
  \midrule
    \MODELNAME{}        & \textbf{19.19x} & \underline{31.20} & \underline{42.07} & \textbf{16.80x} & \underline{28.11} & \underline{37.97} & \textbf{16.85x} & \underline{74.47}  & \underline{78.28} & \textbf{18.24x} & \underline{78.15} & \underline{80.12}  \\
  \bottomrule
  \end{tabular}
  \vspace{-1mm}
  \caption{Open-domain \emph{multi-hop} QA results with Flan-UL2 as the LM $\mathcal{M}$. 
  \textbf{Bold} and \underline{underscore} denote the best and second-best results, respectively.}
  \vspace{-3mm}
  \label{tab-result-multihop}
\end{table*}

%% file: table/tab_result_singlehop.tex
\begin{table}[t]
  \centering
  \small
  \setlength{\tabcolsep}{1.5pt}
  \begin{tabular}{lcccccc}
  \toprule
                        &  \multicolumn{3}{c}{TriviaQA} &  \multicolumn{3}{c}{NQ}      \\
    Method              &   Rate   &   EM    &   F1     &   Rate   &   EM   &   F1     \\
  \midrule
    No documents        &   -      &  49.33  &  54.85   &   -      &  21.99 &  29.38   \\
    Top-5 documents     &   1x     &  62.37  &  70.09   &   1x     &  39.39 &  48.28   \\
  \midrule
    T5                  &   9.80x  &  54.72  &  61.91   &   9.74x  &  30.97 &  38.84   \\
    GPT-3.5             &  15.71x  &  \textbf{62.03}  &  \textbf{69.66}   &  11.33x  &  \textbf{37.12} &  \textbf{46.35}   \\
    Selective Context   &  4.41x   &  52.76  &  59.44   &   4.42x  &  25.51 &  34.05 \\
    LLMLingua           &  4.66x   &  48.24  &  54.58   &   4.62x  &  22.58 &  30.59\\
    RECOMP              &  \underline{16.23x}  &  58.68  &  66.34   &  \underline{11.99x}  &  \underline{37.04} &  \underline{45.47}   \\
  \midrule
    \MODELNAME{}        &  \textbf{29.76x}  &  \underline{59.82}  &  \underline{66.60}   &  \textbf{17.67x}  &  36.40 &  45.00   \\
  \bottomrule
  \end{tabular}  
  \caption{Open-domain \emph{single-hop} QA results with Flan-UL2 as the LM $\mathcal{M}$. 
  }
  \vspace{-2mm}
  \label{tab-result-singlehop}
\end{table}

%% file: text/2_related.tex
\section{Related Work}

Processing and understanding long contexts presents several challenges, including increased inference costs, longer latency, and decreased performance due to redundant and distracting information.
One line of research proposes compressing long contexts into soft prompts that can be used by LMs, such as GIST~\citep{DBLP:conf/nips/Mu0G23}, AutoCompressors~\citep{DBLP:conf/emnlp/ChevalierWAC23}. 
However, these soft prompts are usually tailored to particular tasks and require fine-tuning to get aligned to the representation space of LMs, which severely limits their application scenarios.
Another line of work proposes compressing long contexts into textual summaries, such as LLMLingua~\citep{DBLP:conf/emnlp/JiangWLYQ23}, RECOMP~\citep{DBLP:conf/iclr/XuSC24}, CompAct~\cite{DBLP:conf/emnlp/YoonLHJK24}, and our method belongs to this category.
Compared to soft prompts, this approach yields more interpretable textual summary that can transfer across different LMs, and can be applied to black-box LMs without requiring gradient updates. 
LLMLingua proposes demonstration- and token-level prompt compression methods which leverage a small LM to prune out redundant demonstrations and tokens.
RECOMP distills the summarization ability of extreme-scale proprietary LLMs into an in-house abstractive compressor.
Concurrent to our work, CompAct employs an active strategy to recurrently acquire new information from documents and compress it into a compacted context.

%% file: text/6_conclusion.tex
\section{Conclusion}
This work introduces \MODELNAME{}, a context compressor tailored for document compression to enable multi-hop reasoning with RAG. 
\MODELNAME{} is trained using synthetic data through a pipeline designed to enhance the awareness of multi-hop reasoning, without relying on proprietary LLMs.
Our synthetic data pipeline offers a data-centric approach to constructing high-quality and cost-effective synthetic data for learning context compression. 
Experimental results show that \MODELNAME{} produces more concise summaries while still enabling LMs to show better QA performance than previous compression methods.
\MODELNAME{} also demonstrates competitive QA performance and compression efficiency compared to proprietary LLM GPT-3.5.

\section*{Limitations}
Following the in-distribution setting as used in~\citet{DBLP:conf/iclr/XuSC24}, all compressors (including our proposed \MODELNAME{}) were trained on the training sets of each dataset, and evaluated on either the dev or test sets. 
We conducted a preliminary study on the generalization ability of the compressors to see whether a unified compressor could be trained once on the combined training sets of all datasets used in this work, and then evaluated across all datasets.
The unified compressor is named \UNIMODELNAME{} and the results is shown in Table~\ref{tab-unify} of Appendix~\ref{sec-appendix-results}.
The results indicate that training a unified compressor is promising, as performance improved on two of the datasets. However, it remains challenging, as performance did not improve on the remaining four datasets.
This highlights the need for further optimization and potentially dataset-specific fine-tuning to address the unique characteristics of each dataset and ensure more consistent performance across diverse applications.
This is beyond the scope of this work and will be explored in our future research.

%% file: text/_appendix.tex
\section{Proposition Example} \label{sec-appendix-proposition}
\input{figure/fig_proposition_example}

\section{Prompts} \label{sec-appendix-prompts}
\input{figure/fig_inference_prompt}
\clearpage

\input{figure/fig_generation_multihop_prompt}
\vspace{1.5cm}
\input{figure/fig_decomposition_prompt}
\clearpage

\section{Validation Heuristics} \label{sec-appendix-heuristic}
\input{figure/fig_heuristics}

\section{Implementation Details} \label{sec-appendix-training}
\paragraph{Training Details of Compressor $\mathcal{C}$}  
We train the summarizer using the Adam optimizer, using a batch size of 16 (4 GPUs, with batch size of 4 on each and gradient accumulation steps set to 1), a learning rate of 3e-5 and a constant with warmup learning rate scheduler for 1000 warmup steps with random seed 42. For most times, training for 3 epochs shows the best performance on the development set. Since we are finetuing a T5 model, we also keep the `summarize:' prefix as this shows better results than without it.

\paragraph{Reproduction of the Selective Context baseline} 
In the Selective Context paper~\citep{DBLP:conf/emnlp/0001DGL23}, they choose different LMs to compute self-information for different reader LMs. For example, in their paper they recommend using curie (one variant of gpt-3) for gpt-3.5-turbo and llama2-7b for llama series. We use their default model gpt-2 as the compressor. 
The maximum reduced ratio reported in their paper is 0.8, which equals around 5x of compression. We follow this setting. 
We directly use the `phrase' as the masked lexical unit as it was proved in their paper that this is the optimal choice instead of `token' or `sent'.
\clearpage

\section{Training a Unified Compressor} \label{sec-appendix-results}
\input{table/tab_result_unify}
Following the in-distribution setting as used in~\citet{DBLP:conf/iclr/XuSC24}, all compressors (including our proposed \MODELNAME{}) were trained on the training sets of each dataset, and evaluated on either the dev or test sets. 
We conducted a preliminary study on the generalization ability of the compressors to see whether a unified compressor could be trained once on the combined training sets of all datasets used in this work, and then evaluated across all datasets.
The unified compressor is named \UNIMODELNAME{} and the results is shown in Table~\ref{tab-unify}.
The results indicate that training a unified compressor is promising, as performance improved on two of the datasets. However, it remains challenging, as performance did not improve on the remaining four datasets.
This highlights the need for further optimization and potentially dataset-specific fine-tuning to address the unique characteristics of each dataset and ensure more consistent performance across diverse applications.
This is beyond the scope of this work and will be explored in our future research.

%% file: figure/fig_proposition_example.tex
\begin{figure*}[htb]
    \begin{tcolorbox}
    \textbf{Original Document:} \\
    Prior to restoration work performed between 1990 and 2001, the tower leaned at an angle of 5.5 degrees, but the tower now leans at about 3.99 degrees. This means the top of the Learning Tower of Pisa is displaced horizontally 3.9 meters (12 ft 10 in) from the center.
    \\
    
    \textbf{Decomposed Propositions:}\\
    1. Prior to restoration work performed between 1990 and 2001, the Leaning Tower of Pisa leaned at an angle of 5.5 degrees. \\
    2. The Leaning Tower of Pisa now leans at about 3.99 degrees. \\
    3. The top of the Leaning Tower of Pisa is displaced horizontally 3.9 meters (12 ft 10 in) from the center. 
    \end{tcolorbox}
    \caption{An example of parsing a document into a set of propositions. Atomic proposition expressions can encapsulate distinct factoids in a concise and self-contained natural language format.}
    \label{fig-proposition-example}
\end{figure*}

%% file: figure/fig_inference_prompt.tex
\begin{tcolorbox}[breakable]
\textbf{Inference Prompts for Each Dataset} \\
\\
\textbf{HotpotQA, MultiHop-NQ, MultiHop-TriviaQA:}\\
Which magazine was started first Arthur's Magazine or First for Women?\\
Answer: Arthur's Magazine\\
\\
The Oberoi family is part of a hotel company that has a head office in what city?\\
Answer: Delhi\\
\\
Musician and satirist Allie Goertz wrote a song about the "The Simpsons" character Milhouse, who Matt Groening named after who?\\
Answer: President Richard Nixon\\
\\
What nationality was James Henry Miller's wife?\\
Answer: American\\
\\
Cadmium Chloride is slightly soluble in this chemical, it is also called what?\\
Answer: alcohol\\
\\
\{retrieved\_documents\}\\
\{question\}\\
Answer:\\
\\
\textbf{MuSiQue :}\\
Who was ordered to force a Tibetan assault into the region conquered by Yellow Tiger in the mid-17th century?\\
Answer: Ming general Qu Neng\\
\\
What date was the start of the season of Grey's Anatomy where Eric died?\\
Answer: September 25, 2014\\
\\
When did the publisher of Tetrisphere unveil their new systems?\\
Answer: October 18, 1985\\
\\
Who is the composer of Rhapsody No. 1, named after and inspired by the county where Alfred Seaman was born?\\
Answer: Ralph Vaughan Williams\\
\\
What region is Qaleh Now-e Khaleseh in Mahdi Tajik's birth city located?\\
Answer: Qaleh Now Rural District\\
\\
\{retrieved\_documents\}\\
\{question\}\\
Answer:\\
\\
\textbf{TriviaQA:} \\
Which British politician was the first person to be made an Honorary Citizen of the United States of America? \\ 
Answer: Winston Churchill  \\
\\
Which event of 1962 is the subject of the 2000 film Thirteen Days starring Kevin Costner?  \\
Answer: The Cuban Missile Crisis  \\
\\
Which country hosted the 1968 Summer Olympics?  \\
Answer: Mexico  \\
\\
In which city did the assassination of Martin Luther King?  \\
Answer: MEMPHIS, Tennessee  \\
\\
Which German rye bread is named, according to many reliable sources, from the original meaning `Devil's fart'?  \\
Answer: Pumpernickel\\
\\
\{retrieved\_documents\}\\
\{question\}\\
Answer:\\
\\
\textbf{NQ:}\\
who won a million on deal or no deal\\
Answer: Tomorrow Rodriguez\\
\\
who is the woman washing the car in cool hand luke\\
Answer: Joy Harmon\\
\\
who is the actor that plays ragnar on vikings\\
Answer: Travis Fimmel\\
\\
who said it's better to have loved and lost\\
Answer: Alfred , Lord Tennyson\\
\\
name the first indian woman to be crowned as miss world\\
Answer: Reita Faria\\
\\
\{retrieved\_documents\}\\
\{question\}\\
Answer:\\
\end{tcolorbox}
\captionof{figure}{Inference prompts of Flan-UL2 for HotpotQA, MuSiQue, MultiHop-NQ, MultiHop-TriviaQA, TrivaQA, and NQ respectively. Following RECOMP, we include few-shot in-context examples in the prompt, followed by the retrieved documents (or compressed summary) and the question. The five in-context examples are randomly sampled from their corresponding training sets.
}
\label{fig-inference-prompt}

%% file: figure/fig_generation_multihop_prompt.tex
    \begin{tcolorbox}[breakable]
    \textbf{Multi-hop Question Composition Prompt:} \\
    A multi-hop question requires multiple inferential steps across different pieces of information. Using the provided Wikipedia passages, generate one multi-hop question. Be sure to generate multi-hop questions that are reasonable and factually accurate based on the given articles.\\
\\
Instructions:\\
1. **Find the Connection**: Identify relationships across separate passages. Do not use relations within a single passage. Use bridge entities [S] to connect information.\\
   Example relationships:\\
    - Which continent is [S] located in?\\
    - What is the capital of [S]?\\
    - What is the name of the current head of state in [S]?\\
    - What is the name of the current head of the [S] government?\\
    - Which city did [S] die in?\\
    - Who is [S] married to?\\
    - Which religion is [S] affiliated with?\\
    - What language does [S] speak?\\
    - Which city was [S] born in?\\
    - Which university was [S] educated at?\\
    - Who is [S]’s child?\\
    - What is the country of citizenship of [S]?\\
    - Who performed [S]?\\
    - Who is the employer of [S]?\\
    - Who founded [S]?\\
    - Who is the author of [S]?\\
    - Who was [S] created by?\\
    - Which language was [S] written in?\\
    - What is the official language of [S]?\\
    - Where was [S] founded?\\
    - Which country was [S] created in?\\
    - What kind of work does [S] do?\\
    - What type of music does [S] play?\\
    - What is the original language of [S]?\\
    - Which city did [S] work in?\\
    - What is [S] famous for?\\
    - Which sport is [S] associated with?\\
    - What position does [S] play?\\
    - Who is the head coach of [S]?\\
    - Which city is the headquarter of [S] located in?\\
    - Who is the developer of [S]?\\
    - Who is the chairperson of [S]?\\
    - Who is the chief executive officer of [S]?\\
    - Who is the original broadcaster of [S]?\\
    - Which company is [S] produced by?\\
    - Who is the director of [S]?\\
    - Who is the [S]?\\
\\
2. **Locate Supporting Facts**: Ensure the question involves multiple passages. Label the source passages and the relationship chain.\\
   Example: \\
    - [Passage 1] The continent of the country of [S2] is located in [S1]. [Passage 2] The author of [S3] is [S2]. [Combine] What continent is the country of citizenship of the author of [S] located in?\\
    - [Passage 1] The nationality of the author of [S2] is [S1]. [Passage 2] The novel [S2] was adapted into [S3]. [Combine] What is the nationality of the author of the novel that was adapted into [S]?\\
    - [Passage 1] The child of the [S2] was educated in [S1]. [Passage 2] [S2] is the chairperson of [S3]. [Combine] Which university was the child of the chairperson of [S] educated?\\
    - [Passage 1] [S2] speaks the language [S1]. [Passage 2] [S3] was developed by [S2]. [Combine] What language does the developer of [S] speak?\\
\\
3. **Question Construction Rules**:\\
   - Do not use more than one what/why/how/...\\
     - Not allowed: What kind of work does [S] do and who is [S]’s child?\\
     - Allowed: What kind of work does the child of [S] do?\\
   - Do not include the intermediate entity in the question.\\
     - Not allowed: What is the date of independence for [S1], which was predominantly populated by [S2]?\\
     - Allowed: What is the date of independence for the country that was predominantly populated by [S]?\\
\\
4. **Generate Answer**: Provide an answer based on the passages.\\
\\
If no meaningful multi-hop question can be generated, reply with "Sorry, I cannot generate any multi-hop question based on the provided passages."\\
\\
Examples:\\
\\
Passages:\\
1. James Henry Miller (25 January 1915 - 22 October 1989), better known by his stage name Ewan MacColl, was an English folk singer, songwriter, communist, labour activist, actor, poet, playwright and record producer.\\
2. Margaret "Peggy" Seeger (born June 17, 1935) is an American folksinger. She is also well known in Britain, where she has lived for more than 30 years, and was married to the singer and songwriter Ewan MacColl until his death in 1989.\\
Supporting Facts:\\
1. [Passage 1] James Henry Miller (25 January 1915 - 22 October 1989), better known by his stage name Ewan MacColl, was an English folk singer, songwriter, communist, labour activist, actor, poet, playwright and record producer. \\
2. [Passage 2] Margaret "Peggy" Seeger is also well known in Britain, where she has lived for more than 30 years, and was married to the singer and songwriter Ewan MacColl until his death in 1989. \\
3. [Passage 2] Margaret "Peggy" Seeger (born June 17, 1935) is an American folksinger.\\
Relationship Chain:\\
James Henry Miller is Ewan MacColl. Ewan MacColl is married to Margaret. Margaret is American. So, the nationality of James Henry Miller's wife is American.\\
Multihop Question: \\
What nationality was James Henry Miller's wife?\\
Answer: \\
American\\
\\
\\
Passages:\\
1. The Oberoi family is an Indian family that is famous for its involvement in hotels, namely through The Oberoi Group.\\
2. The Oberoi Group is a hotel company with its head office in Delhi. Founded in 1934, the company owns and/or operates 30+ luxury hotels and two river cruise ships in six countries, primarily under its Oberoi Hotels \& Resorts and Trident Hotels brands.\\
Supporting Facts:\\
1. [Passage 1] The Oberoi family is an Indian family that is famous for its involvement in hotels, namely through The Oberoi Group.\\
2. [Passage 2] The Oberoi Group is a hotel company with its head office in Delhi. [Relation: The Oberoi Group's head office is in Delhi.]\\
Relationship Chain:\\
The Oberoi family involve the hotel industry through the Oberoi Group. The Oberoi Group's head office is in Delhi. So the Oberoi family is part of a hotel company that has a head office in Delhi.\\
Multihop Question:\\
The Oberoi family is part of a hotel company that has a head office in what city?\\
Answer:\\
Delhi\\
\\
Now, it's your turn. Ensure there's only one what/why/how/... in your question and that the relationship chain spans multiple passages.\\
Passages:\\
\{given\_doc\}\\
    \\

    \end{tcolorbox}
    \captionof{figure}{The prompt used to compose multi-hop questions given multiple documents. We utilize the most common relationships identified by \citet{DBLP:conf/emnlp/ZhongWMPC23} to aid in discovering connections between separate documents. The few-shot examples are sampled from HotpotQA and labeled by us.}
    \label{fig-generation-multihop-prompt}

%% file: figure/fig_decomposition_prompt.tex
\begin{tcolorbox}[breakable]
\textbf{Multi-hop Question Decomposition Prompt:} \\
A multi-hop question requires multiple inferential steps or accessing information from different sources. Given a multi-hop question and its context, your task is to decompose it into single-hop questions. Be sure to generate single-hop questions that are reasonable and factually accurate. Ensure that the decomposition remains true to the original multi-hop question and does not introduce any inaccuracies or hallucinations. You should decompose the multi-hop question based merely on the multi-hop question, and the context is only for the answer of the single-hop questions.\\
\\
Here are some instructions:\\
1. Find the bridge entity: A bridge entity is the key element linking the parts of the multi-hop question. It should be the answer to one single-hop question and appear in the description of the other single-hop question. **Important: The bridge entity is not the answer to the multi-hop question and will not appear in the multi-hop question. Ensure you do not use the bridge entity as the answer to the multi-hop question.**\\
2. Recover questions: After identifying the bridge entity, decompose the multi-hop question into two single-hop questions. Ensure one question can be answered with the bridge entity, while the other question includes the bridge entity in its description.\\
\\
Examples:\\
Question: What is the independence date of the country where the majority of the population is composed of Ambundu, Ovimbundu, and Bakongo peoples?\\
Answer: 11 November 1975\\
Context 1: It is thus reasonable to talk of Angola as a defined territorial entity from this point onwards. In 1961, the FNLA and the MPLA, based in neighbouring countries, began a guerrilla campaign against Portuguese rule on several fronts. The Portuguese Colonial War, which included the Angolan War of Independence, lasted until the Portuguese regime's overthrow in 1974 through a leftist military coup in Lisbon. When the timeline for independence became known, most of the roughly 500,000 ethnic Portuguese Angolans fled the territory during the weeks before or after that deadline. Portugal left behind a newly independent country whose population was mainly composed by Ambundu, Ovimbundu, and Bakongo peoples.\\
Context 2: This was ratified by the Alvor Agreement later that month, which called for general elections and set the country's independence date for 11 November 1975. All three factions, however, followed up on the ceasefire by taking advantage of the gradual Portuguese withdrawal to seize various strategic positions, acquire more arms, and enlarge their militant forces. The rapid influx of weapons from numerous external sources, especially the Soviet Union and the United States, as well as the escalation of tensions between the nationalist parties, fueled a new outbreak of hostilities. With tacit American and Zairean support the FNLA began massing large numbers of troops in northern Angola in an attempt to gain military superiority.\\
\\
Bridge Entity: Angola\\
Recovered Questions:\\
1. Question: What is the independence date of Angola?\\
   Answer: 11 November 1975\\
2. Question: What country has a majority population of Ambundu, Ovimbundu, and Bakongo peoples?\\
   Answer: Angola\\
\\
\\
Question: What themes are explored in the work that inspired "2001: A Space Odyssey"?\\
Answer: Human evolution\\
Context 1: Since its premiere, "2001: A Space Odyssey" has been analyzed and interpreted by professional critics and theorists, amateur writers, and science fiction fans. Peter Kramer in his monograph for BFI analyzing the film summarized the diverse interpretations as ranging from those who saw it as darkly apocalyptic in tone to those who saw it as an optimistic reappraisal of the hopes of mankind and humanity. Questions about "2001" range from uncertainty about its implications for humanity's origins and destiny in the universe to interpreting elements of the film's more enigmatic scenes, such as the meaning of the monolith, or the fate of astronaut David Bowman.\\
Context 2: "2001: A Space Odyssey" (film) is a 1968 epic science fiction film produced and directed by Stanley Kubrick. The screenplay was written by Kubrick and Arthur C. Clarke and was inspired by Clarke's short story "The Sentinel". Written concurrently with the screenplay, a novel was published soon after the film was released. The film, which follows a voyage to Jupiter with the sentient computer HAL after the discovery of a mysterious black monolith affecting human evolution, deals with themes of existentialism, human evolution, technology, artificial intelligence, and the possibility of extraterrestrial life. The film is noted for its scientifically accurate depiction of spaceflight, pioneering special effects, and ambiguous imagery.\\
\\
Bridge Entity: "The Sentinel"\\
Recovered Questions:\\
1. Question: What themes are explored in "The Sentinel"?\\
   Answer: Human evolution\\
2. Question: What work inspired "2001: A Space Odyssey"?\\
   Answer: "The Sentinel"\\
\\
Now, it's your turn.\\
Question: \{question\}\\
Answer: \{answer\}\\
\{context\}\\
\\

\end{tcolorbox}
\captionof{figure}{The prompt used to decompose multi-hop questions into a set of single-hop questions, their corresponding answers and the bridge entities.}
\label{fig-decomposition-prompt}

%% file: figure/fig_heuristics.tex
\begin{figure*}[htb]
    \begin{tcolorbox}
    Given a multi-hop question and its ground truth answer, we prompt Llama3-70B-Instruct under temperature of 0 to decompose it into single-hop questions with bridge entities. \\The following rules can be used to validate if the `bridge' type of questions are multi-hop. We leave out the `comparison' type in this case.   
    \\
    \textbf{Heuristic rules:} 
    \begin{itemize}
        \item The ground truth should not be one of the bridge entities.
        \item The bridge entities should not appear in the multi-hop question.
        \item A multi-hop reasoning path can be found in the decomposed single-hop questions, e.g. one bridge entity should both appear in single-hop question 1 and be the answer to single-hop question 2. The end of this reasoning path should be the ground truth answer.
        \item All the decomposed single-hop questions should be correctly answered by prepending one of the propositions with an improved likelihood.
        \item The propositions must come from different documents.
    \end{itemize}
    \end{tcolorbox}
    \caption{Heuristic rules used to validate whether a question is multi-hop.}
    \label{fig-heuristics}
\end{figure*}

%% file: table/tab_result_unify.tex
\begin{table}[!hbt]
  \centering
  \resizebox{0.48\linewidth}{!}{  
  \begin{tabular}{llccc}
  \toprule
                                     & Method          & Rate   &   EM  &   F1   \\
  \midrule
  \multirow{2}{*}{Multihop-TriviaQA} & \MODELNAME{}    & 18.24x & 78.15 & 80.12  \\
                                     & \UNIMODELNAME{} & 18.98x & 80.91 & 82.60  \\
  \midrule
  \multirow{2}{*}{Multihop-NQ}       & \MODELNAME{}    & 16.85x & 74.47 & 78.28  \\
                                     & \UNIMODELNAME{} & 17.04x & 75.74 & 78.65  \\
  \midrule
  \multirow{2}{*}{HotpotQA}          & \MODELNAME{}    & 19.19x & 31.20 & 42.07  \\
                                     & \UNIMODELNAME{} & 21.99x & 29.40 & 38.70  \\
  \midrule
  \multirow{2}{*}{MuSiQue}           & \MODELNAME{}    & 16.80x & 28.11 & 37.97  \\
                                     & \UNIMODELNAME{} & 17.50x & 27.16 & 37.25  \\
  \midrule
  \multirow{2}{*}{TriviaQA}          & \MODELNAME{}    & 29.76x & 59.82 & 66.60  \\
                                     & \UNIMODELNAME{} & 29.76x & 59.77 & 66.60  \\
  \midrule
  \multirow{2}{*}{NQ}                & \MODELNAME{}    & 17.67x & 36.40 & 45.00  \\
                                     & \UNIMODELNAME{} & 17.45x & 36.32 & 45.44  \\
  \bottomrule
  \end{tabular}  
  }
  \caption{The comparison of \MODELNAME{} and \UNIMODELNAME{} on all six datasets used in this work.}
  \vspace{-2mm}
  \label{tab-unify}
\end{table}